\documentclass[graybox]{svmult}

\usepackage{mathptmx}       
\usepackage{helvet}         
\usepackage{courier}        
\usepackage{type1cm}        
\usepackage{makeidx}         
\usepackage{graphicx}        
\usepackage{multicol}        
\usepackage[bottom]{footmisc}

\usepackage{hyperref}
\usepackage{times}
\usepackage{helvet}
\usepackage{courier}
\usepackage{amssymb}
\usepackage{latexsym}
\usepackage{amsmath}
\usepackage{multirow}
\usepackage{url}
\usepackage{algorithm}
\usepackage{algpseudocode}
\usepackage{amsmath}
\usepackage{rotating}
\usepackage{epsf}
\usepackage{enumitem}

\makeindex             

\begin{document}

\title*{{\it SimpleDS}: A Simple Deep Reinforcement Learning Dialogue System}
\author{Heriberto Cuay\'ahuitl$^1$}
\institute{$^1$Computer Science at Heriot-Watt University, Edinburgh, Scotland, \email{hc213@hw.ac.uk}}
%
%
\maketitle

\abstract*{This paper presents {\it SimpleDS}, a simple and publicly available dialogue system trained with deep reinforcement learning. 
In contrast to previous reinforcement learning dialogue systems, this system avoids manual feature engineering by performing action selection directly from raw text of the last system and (noisy) user responses. Our initial results, in the restaurant domain, show that it is indeed possible to induce reasonable dialogue behaviour with an approach that aims for high levels of automation in dialogue control for intelligent interactive agents.}

\abstract{
This paper presents {\it SimpleDS}, a simple and publicly available dialogue system trained with deep reinforcement learning. In contrast to previous reinforcement learning dialogue systems, this system avoids manual  feature engineering by performing action selection directly from raw text of the last system and (noisy) user responses. Our initial results, in the restaurant domain, report that it is indeed possible to induce reasonable behaviours with such an approach that aims for higher levels of automation in dialogue control for intelligent interactive agents.
}

\section{Introduction}
Almost two decades ago, the (spoken) dialogue systems community adopted the Reinforcement Learning (RL) paradigm since it offered the possibility to treat dialogue design as an optimisation problem, and because RL-based systems can improve their performance over time with experience. Although a large number of methods have been proposed for training (spoken) dialogue systems using RL, the question of ``How to train dialogue policies in an efficient, scalable and effective way across domains?'' still remains as an open problem. One limitation of current approaches is the fact that RL-based dialogue systems still require high-levels of human intervention (from system developers), as opposed to automating the dialogue design. Training a system of this kind requires a system developer to provide a set of features to describe the dialogue state, a set of actions to control the interaction, and a performance function to reward or penalise the action-selection process. All of these elements have to be carefully engineered in order to learn a good dialogue policy (or policies). This suggests that one way of advancing the state-of-the-art in this field is by reducing the amount of human intervention in the dialogue design process through higher degrees of automation, i.e. by moving towards truly autonomous learning.

Recent advances in machine learning have proposed machine learning methods as a way to reduce human intervention in the creation of intelligent agents. In particular, the field of Deep Reinforcement Learning (DRL) targets feature learning and policy learning simultaneously---which reduces the effort in feature engineering \cite{mnih-atari-2013}. This is relevant because the vast majority of previous RL-based dialogue systems make use of carefully engineered features to represent the dialogue state \cite{PaekP08}. 

Motivated by the advantages of DRL methods over traditional RL methods, in this paper we present an extended dialogue system, recently applied to strategic dialogue management\cite{CuayahuitlEtAl2015nips}, that makes use of raw noisy text---without any engineered features to represent the dialogue state. By using this representation, the dialogue system does not require a Spoken Language Understanding (SLU) component. We bypass SLU by learning
 dialogue policies directly from (simulated) speech recognition outputs. The rest of the paper describes a proof of concept system  which is trained based on this idea. 

\section{Deep Reinforcement Learning for Dialogue Control}
\label{DRL}
A Reinforcement Learning (RL) agent learns its behaviour
from interaction with an environment and the physical or virtual agents within it, where situations are mapped to actions by maximising a long-term reward signal \cite{Szepesvari:2010}. An RL agent is typically  characterised by: (i) a finite or infinite set of states $S=\{s_i\}$; (ii) a finite or infinite set of actions $A=\{a_j\}$; (iii) a state transition function $T(s,a,s')$ that specifies the next state $s'$ given the current state $s$ and action $a$; (iv) a reward function $R(s,a,s')$ that specifies the reward given to the agent for choosing action $a$ in state $s$ and transitioning to state $s'$; and (v) a policy $\pi:S \rightarrow A$ that defines a mapping from states to actions. 
The goal of an RL agent is to select actions by maximising its cumulative discounted reward defined as $Q^*(s,a)=\max_\pi \mathbb{E}[r_t+\gamma r_{t+1}+\gamma^2 r_{t+1}+...|s_t=s,a_t=a,\pi]$, 
where function $Q^*$ represents the maximum sum of rewards $r_t$  discounted by factor $\gamma$ at each time step. While the RL  agent takes actions with probability $Pr(a|s)$ during training, it takes the best actions $\max_a Pr(a|s)$ at test time.

To induce the $Q$ function above we use Deep Reinforcement Learning as in \cite{mnih-atari-2013}, which approximates $Q^*$ using a multilayer convolutional neural network. The $Q$ function of a DRL agent is parameterised as $Q(s,a;\theta_i)$, where $\theta_i$ are the parameters (weights) of the neural net at iteration $i$. More specifically, training a DRL agent requires a dataset of experiences $D_t=\{e_1,...e_t\}$ (also referred to as `experience replay memory'), where every experience is described as a tuple  $e_t=(s_t,a_t,r_t,s_{t+1})$.
The $Q$ function can be induced by applying Q-learning updates over minibatches of experience $MB=\{(s,a,r,s')\sim U(D)\}$ drawn uniformly at random from dataset $D$. A Q-learning update at iteration $i$ is thus defined as the loss function $L_i(\theta_i)=\mathbb{E}_{MB} \left[ (r+\gamma \max_{a'} Q(s',a';\overline{\theta}_i)-Q(s,a;\theta_i))^2 \right]$, 
where $\theta_i$ are the parameters of the neural net at iteration $i$, and $\overline{\theta}_i$ are the target parameters of the neural net at iteration $i$. The latter are only updated every $C$ steps. This process is implemented in the learning algorithm {\it Deep Q-Learning with Experience Replay} described in \cite{mnih-atari-2013}. 

\section{The {\it SimpleDS} Dialogue System}

\begin{figure*}[th]
  \begin{center}
    \includegraphics[width=0.78\textwidth]{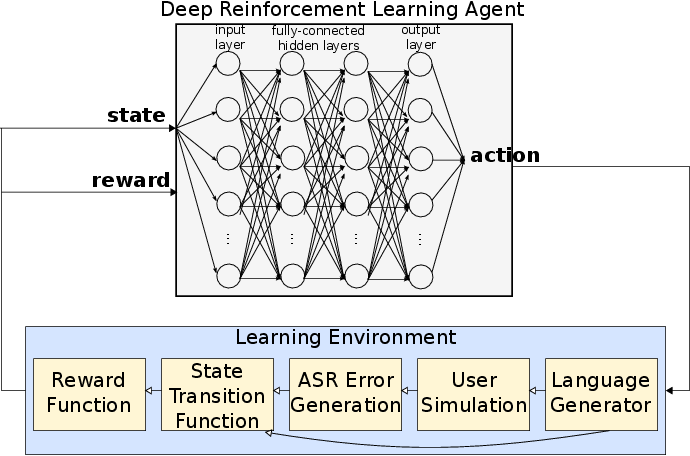}
\caption{\label{integratedSystem} High-level architecture of the {\it SimpleDS} dialogue system--see text for details.}
  \end{center}
\end{figure*}

Figure~\ref{integratedSystem} shows a high-level diagram of the {\it SimpleDS} dialogue system. At the bottom, the learning environment receives an action (dialogue act) and outputs the next environment state and numerical reward. To do that, the environment first generates the word sequence of the last system action, the user simulator generates a word sequence as a response to that action, and the user response is distorted given some noise level and word-level confidence scores. Based on the system's verbalisation and noisy user response, the next dialogue state and reward are calculated and given as a result of having executed the given action. At the top of the diagram, a Deep Reinforcement Learning (DRL) agent receives the state and reward, updates its policy during learning, and outputs an action according to its learnt policy.

This system runs under a client-server architecture, where  the environment acts as the {\it server} and the learning agent acts as the {\it client}. They communicate by exchanging messages, where the client tells the server the action to execute, and the server tells the client the dialogue state and reward observed. The {\it SimpleDS} learning agent is based on the ConvNetJS tool \cite{ConvNetJS}, which implements the algorithm `Deep Q-Learning with experience replay' proposed by \cite{mnih-atari-2013}. We extended this tool to support multi-threaded and client-server processing with constrained search spaces.\footnote{The code of {\it SimpleDS} is available at \url{https://github.com/cuayahuitl/SimpleDS}} 

The {\it state space} includes up to 100 word-based features depending on the vocabulary 
of the {\it SimpleDS} agent in the restaurant domain. The initial release of {\it SimpleDS} provides support for English, German and Spanish. While words derived from system responses are treated as binary variables (i.e. word present or absent), the words derived from noisy user responses can be seen as continuous variables by taking confidence scores into account. Since we use a single variable per word, user features override system ones in case of overlaps. 

The {\it action space} includes 35 dialogue acts in the Restaurant domain\footnote{Actions: Salutation(greeting), Request(hmihy), Request(food), Request(price), Request(area), Request(food, price), Request(food, area), Request(price, area), Request(food, price, area), AskFor(more), Apology(food), Apology(price), Apology(area), Apology(food, price), Apology(food, area), Apology(price, area), Apology(food, price, area), ExpConfirm(food), ExpConfirm(price), ExpConfirm(area), ExpConfirm(food, price), ExpConfirm(food, area), ExpConfirm(price, area), ExpConfirm(food, price, area), ImpConfirm(food), ImpConfirm(price), ImpConfirm(area), ImpConfirm(food,price), ImpConfirm(food, area), ImpConfirm(price, area), ImpConfirm(food, price, area), Retrieve(info), Provide(unknown), Provide(known), Salutation(closing).}. They include 2 salutations, 9 requests, 7 apologies, 7 explicit confirmations, 7 implicit confirmations, 1 retrieve information, and 2 provide information. Rather than learning with whole action sets, $SimpleDS$ supports learning from constrained actions by applying Q-learning learning updates only on the set of valid actions. The constrained actions come from the most likely actions (e.g. $Pr(a|s)>0.01$) with probabilities derived from a Naive Bayes classifier trained from example dialogues. The latter are motivated by the fact that a new system does not have training data apart from a small number of demonstration dialogues. In addition to the most probable data-like actions, the constrained action set is extended with the legitimate requests, apologies and confirmations in state $s$. The fact that constrained actions are data-driven and driven by  application independent heuristics facilitates its usage across domains. 

The {\it state transition function} is based on a numerical vector representing the last system and user responses. The former are straightforward, 0 if absent and 1 if present. The latter correspond to the confidence level [0..1] of noisy user responses. Given that {\it SimpleDS} targets a simple and extensible dialogue system, it uses templates for language generation, a rule-based user simulator, and confidence scores generated uniformly at random (words with scores under a threshold were distorted). 

The {\it reward function} is motivated by the fact that human-machine dialogues should confirm the information required and that interactions should be human-like. It is defined as $R(s,a,s')=(CR \times w)+(DR \times (1-w))-DL$, where $CR$ is the number of positively confirmed slots divided by the slots to confirm; $w$ is a weight over the confirmation reward (CR), we used $w$=0.5; $DR$ is a data-like probability of having observed action $a$ in state $s$, and $DL$ is used to encourage efficient interactions, we used $DL$=0.1. The $DR$ scores are derived from the same statistical classifier above, which allows us to do statistical inference over actions given states ($Pr(a|s)$). 


The {\it model architecture} consists of a fully-connected multilayer neural net with up to 100 nodes in the input layer (depending on the vocabulary), 40 nodes in the first hidden layer, 40 nodes in the second hidden layer, and 35 nodes (action set) in the output layer. The hidden layers use Rectified Linear Units to normalise their weights \cite{NairH10}. Finally, the learning parameters are as follows: experience replay size=10000, discount factor=0.7, minimum epsilon=0.01, batch size=32, learning steps=20000. A comprehensive analysis comparing multiple state representations, action sets, reward functions and learning parameters is left as future work.

\begin{figure}[t]
  \begin{center}
    \includegraphics[width=0.69\textwidth]{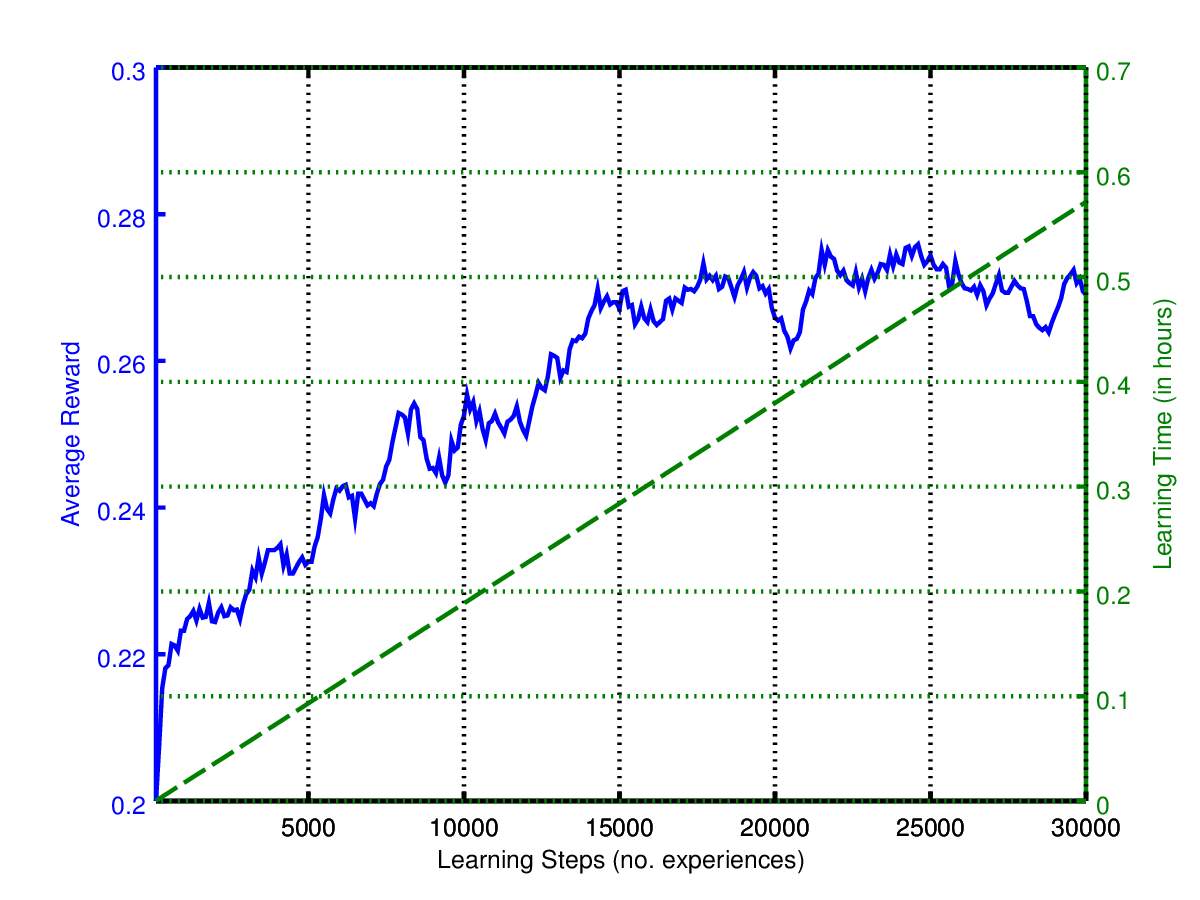}
\caption{\label{LearningCurves} Learning curve of the {\it SimpleDS} Deep Reinforcement Learning agent--see text for details.}
  \end{center}
\end{figure}


Figure~\ref{LearningCurves} shows the learning curve of a {\it SimpleDS} agent using 3000 simulated dialogues. This agent uses a smaller set of actions per state (between 4 and 5 actions) rather than the whole action set per state---according to the  application-independent heuristics mentioned in the previous Section. 
This reduction has the advantage that policies can be learnt quicker, that more sensible dialogues can potentially be learnt, and that it is inherent that some domains make use of legitimate actions during the interaction. In the case of more complex systems, with higher amounts of features and actions, learning with valid actions (rather than all actions) can make a huge difference in terms of computational resources and learning time.  The quality of the learnt policies will depend on the learning environment and given constraints. 

Table~\ref{exampleDialogue} shows an example dialogue of the learnt policy with user inputs derived from simulated speech recognition results. Our initial tests suggest that reasonable interactions can be generated using the proposed learning approach. {\it SimpleDS} has been demonstrated on a mobile App, and a human evaluation is left as future work. 

\begin{table*}[t]
\scriptsize
\centering
\begin{tabular}{|l|l|l|}
\hline
\bf Environment State & \bf Action & \bf Verbalisation \\
\hline 

{\tiny 0,0,0,0,0,0,0,0,0,0,0,0,0,0,0,0,0,0,0,0,0,0,0,0,0,0,0,0,0,0,0,0,0,0,0,0,0,0,0,0,0,} & Salutation & \\
{\tiny 0,0,0,0,0,0,0,0,0,0,0,0,0,0,0,0,0,0,0,0,0,0,0,0,0,0,0,0,0,0,0,0,0,0,0,0,0,0,0,0,0 } & (greeting) & Hello!\\
\hline 
{\tiny 0,0,0,0,0,0,0,0,0,0,0,0,0,0,0,0,0,0,0,0,0,0,0,0,0,0,1,0,0,0,0,0,0,0,0,0,0,0,0,0,0,} & Request & What type of food, price range,\\
{\tiny 0,0,0,0,0,0,0,0,0,0,0,0,0,0,0,0,0,0,0,0,0,0,0,0,0,0,0,0,0,0,0,0,0,0,0,0,0,0,0,0,0 } & (food,price,area) &  and area are you looking for?\\
&  & [reasonably priced mexican \\
&  & ...food in the east of town]\\
\hline 
{\tiny 0,0,0,1,0,1,1,0,0,0,0,0,0,0,0,0,0,0,0.53,0,0,0,0,0.63,1,0,0,0,0,0,0,0.83,0,0,0,0,0,0,0,0,1,0,0,} &ImpConfirm & Okay, reasonably priced \\
{\tiny 0,0.78,0,0,0,0,0.31,0,0,0,1,0.36,1,0.25,0,0,0,0,0,0,0,0,0,0.57,0,0,0.82,1,0,0,0,0,1,0,0,0,0,0,1} & (food,price,area) & mexican food in the east.\\
\hline 
{\tiny 0,0,0,0,0,0,0,0,0,0,0,0,0,0,0,0,0,0,1,0,0,0,0,1,0,0,0,0,0,0,0,1,0,0,0,0,0,0,0,0,0,} & Retrieve & \\
{\tiny 0,0,0,1,0,0,0,0,0,1,0,0,0,1,0,1,0,0,0,0,0,0,0,0,0,1,0,0,0,0,0,0,0,0,0,0,0,0,0,0,0} & (info) & Let me see.\\
\hline  
{\tiny 0,0,0,0,0,0,0,0,0,0,0,0,0,0,0,0,0,0,0,0,0,0,0,0,0,0,0,0,0,0,0,0,0,0,0,0,0,1,0,0,0,} & Provide & Restaurant X is an excellent \\
{\tiny 0,0,1,0,0,0,0,0,0,0,0,0,0,0,0,0,0,0,1,0,0,0,0,0,0,0,0,0,0,0,0,0,0,0,0,0,0,0,0,0,0} & (known) & choice. It is located in Y.\\
\hline  
{\tiny 1,0,0,0,0,0,0,0,0,0,0,0,0,0,0,0,0,1,0,0,0,0,1,0,0,0,0,0,0,1,1,0,0,0,0,0,0,0,0,0,0,} & AskFor & \\
{\tiny 0,1,0,0,0,0,0,0,0,0,0,1,0,0,0,0,0,0,0,0,0,1,0,0,0,0,1,0,0,0,1,0,0,0,0,1,0,0,0,0,0} & (more) & Anything else?\\
& & [no]\\
\hline 
{\tiny 0,0,0,0,1,0,0,0,0,0,0,0,0,0,0,0,0,0,0,1,0,0,0,0,0,0,0,0,0,0,0,0,0,0,0,0,0,0,0,0,0,0,} & Salutation & \\
{\tiny 0,0,0,0,0.72,0,0,0,0,0,0,0,0,0,0,0,0,0,0,0,0,0,0,0,0,0,0,0,0,0,0,0,0,0,0,0,0,0,0,0} & (closing) & Okay, talk to you soon. Bye!\\ 
\hline
{\tiny 0,0,0,0,0,0,0,0,1,0,0,0,0,0,0,0,0,0,0,0,0,0,0,0,0,0,0,0,0,0,0,0,0,0,0,0,0,0,0,0,0,} & & \\
{\tiny 0,0,0,0,0,0,0,0,0,1,0,0,0,0,0,0,0,0,0,0,1,0,0,1,0,0,0,1,0,0,0,0,0,0,0,0,0,0,0,0,1} & & \\
\hline
\end{tabular}
\caption{Example dialogue using the policy from Fig.2, where states are numerical representations of the last system and noisy user inputs, actions are dialogue acts, and user resposes are in brackets}\label{exampleDialogue}
\end{table*}

\section{Summary}
We describe a publicly available dialogue system motivated by the idea that future dialogue systems should be trained with almost no intervention from system developers. In contrast to previous reinforcement learning dialogue systems, {\it SimpleDS} selects dialogue actions directly from raw (noisy) text of the last system and user responses. 
It remains to be demonstrated how far one can go with such an approach. Future work includes to (a) compare different model architectures, training parameters and reward functions; (b) extend or improve the abilities of the proposed dialogue system; (c) train deep learning agents in other (larger scale) domains \cite{CuayahuitlRLS10,CuayahuitlD11,CuayahuitlKD14}; (d) evaluate end-to-end systems with real users; (e) compare or combine different types of neural nets \cite{SainathVSS15}; and (e) perform fast learning based on parallel computing.

\section*{Acknowledgments}
Funding from the European Research Council (ERC) project ``STAC: Strategic Conversation'' no. 269427 is gratefully acknowledged. 

\bibliographystyle{splncs}
\bibliography{hc-iwsds2016}

\begin{thebibliography}{10}

\bibitem{mnih-atari-2013}
Mnih, V., Kavukcuoglu, K., Silver, D., Graves, A., Antonoglou, I., Wierstra,
  D., Riedmiller, M.:
\newblock Playing atari with deep reinforcement learning.
\newblock In: NIPS Deep Learning Workshop.
\newblock (2013)

\bibitem{PaekP08}
Paek, T., Pieraccini, R.:
\newblock Automating spoken dialogue management design using machine learning:
  An industry perspective.
\newblock Speech Communication \textbf{50}(8-9) (2008)

\bibitem{CuayahuitlEtAl2015nips}
Cuay\'ahuitl, H., Keizer, S., Lemon, O.:
\newblock Strategic dialogue management via deep reinforcement learning.
\newblock In: NIPS Deep Reinforcement Learning Workshop.
\newblock (2015)

\bibitem{Szepesvari:2010}
Szepesv\'ari, C.:
\newblock Algorithms for Reinforcement Learning.
\newblock Morgan and Claypool Pub. (2010)

\bibitem{ConvNetJS}
Karpathy, A.:
\newblock {ConvNetJS}: A javascript library for training deep learning models.
\newblock https://github.com/karpathy/convnetjs (2015)

\bibitem{NairH10}
Nair, V., Hinton, G.E.:
\newblock Rectified linear units improve restricted boltzmann machines.
\newblock In: ICML. (2010)

\bibitem{CuayahuitlRLS10}
Cuay{\'{a}}huitl, H., Renals, S., Lemon, O., Shimodaira, H.:
\newblock Evaluation of a hierarchical reinforcement learning spoken dialogue
  system.
\newblock Computer Speech {\&} Language \textbf{24}(2) (2010)

\bibitem{CuayahuitlD11}
Cuay{\'{a}}huitl, H., Dethlefs, N.:
\newblock Spatially-aware dialogue control using hierarchical reinforcement
  learning.
\newblock {TSLP} \textbf{7}(3) (2011)

\bibitem{CuayahuitlKD14}
Cuay{\'{a}}huitl, H., Kruijff{-}Korbayov{\'{a}}, I., Dethlefs, N.:
\newblock Nonstrict hierarchical reinforcement learning for interactive systems
  and robots.
\newblock TiiS \textbf{4}(3) (2014)

\bibitem{SainathVSS15}
Sainath, T.N., Vinyals, O., Senior, A.W., Sak, H.:
\newblock Convolutional, long short-term memory, fully connected deep neural
  networks.
\newblock In: ICASSP. (2015)

\end{thebibliography}
\end{document}